# Bio-Inspired Foveated Technique for Augmented-Range Vehicle Detection Using Deep Neural Networks

Pedro Azevedo, Sabrina S. Panceri, Rânik Guidolini, Vinicius B. Cardoso,
Claudine Badue, Thiago Oliveira-Santos and Alberto F. De Souza, *Senior Member*, *IEEE*
Departamento de Informática
Universidade Federal do Espírito Santo
Vitória, Brasil
{pedro, sabrina.panceri, ranik, vinicius, claudine, todsantos, alberto}@lcad.inf.ufes.br

*Abstract*— We propose a bio-inspired foveated technique to detect cars in a long range camera view using a deep convolutional neural network (DCNN) for the IARA self-driving car. The DCNN receives as input (i) an image, which is captured by a camera installed on IARA's roof; and (ii) crops of the image, which are centered in the waypoints computed by IARA's path planner and whose sizes increase with the distance from IARA. We employ an overlap filter to discard detections of the same car in different crops of the same image based on the percentage of overlap of detections' bounding boxes. We evaluated the performance of the proposed augmented-range vehicle detection system (ARVDS) using the hardware and software infrastructure available in the IARA self-driving car. Using IARA, we captured thousands of images of real traffic situations containing cars in a long range. Experimental results show that ARVDS increases the Average Precision (AP) of long range car detection from 29.51% (using a single whole image) to 63.15%.

## I. INTRODUCTION

Self-driving cars that operate in public roads must be aware of other vehicles around them. Many techniques have being proposed for detecting such neighboring vehicles and these may be roughly divided into RADAR [1]-[3], LiDAR [4]-[6] and camera [7]-[9] based techniques. RADAR and LiDAR based techniques are perhaps the currently most advanced, but RADAR and LiDAR sensors are still expensive. Cameras, on the other hand, are not expensive and, although camera based techniques require a significant amount of processing power for image analysis, processing power might already be available in the self-driving car for other required functionalities. Nevertheless, even though there are currently powerful deep neural networks for vehicles detection in images [10]-[12], they cannot detect them at the required range for many application scenarios.

Humans can detect vehicles at a considerable range using sight. Different from current object detection systems based on deep neural networks, which examine images with throughout uniform resolution, the human biological vision system (and that of many other mammals) has log-polar resolution (resolution is far greater in the fovea [13] [14]). When driving, we continuously move our eyes in order to look at the points in the scenario from where we expect other vehicles might come, i.e., in order to bring such points to our foveae. Inspired by these properties of the human biological vision system, in this paper, we present an investigation of the advantages of using a foveated technique for augmented-range vehicle detection using deep neural networks.

Cheap cameras available nowadays have a considerable high resolution; 1920×1080-pixel resolution is common, for instance. The images captured by them must be rescaled to much lower resolution to feed the deep neural networks used for object-detection, in which an input size of about 416×416-pixel is common in high frame rate object detection systems [10]-[12]. This rescaling makes far objects unrecognizable, though. One solution to this problem is to split the images in small parts (image crops) and give them as input to the neural network, but this would require significant processing power, in addition to the necessity of handling objects in the border of the images' crops. However, if the neural system knew where to look at in the images (as we do when driving), it could select a few crops of the images that would contain the points of required attention (see Fig. 1).

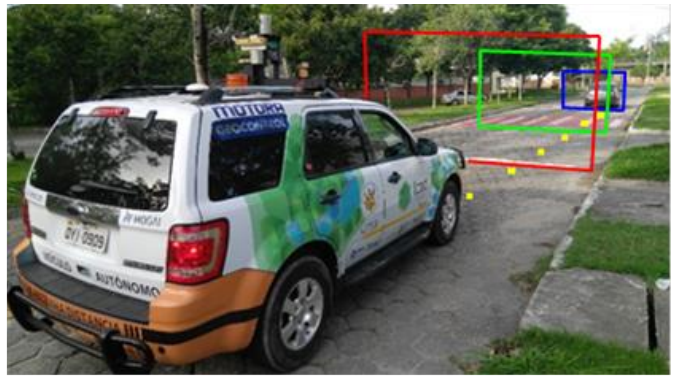

Fig. 1 - The proposed augmented-range vehicle detection system. Red, green and blue rectangles represent image crops used as input in the deep neural network. Yellow dots represent waypoints along the path planned by the self-driving car.

This study was financed in part by Coordenação de Aperfeiçoamento de Pessoal de Nível Superior – Brasil (CAPES) – Finance Code 001; Conselho Nacional de Desenvolvimento Científico e Tecnológico - Brasil (CNPq) - grants 311120/2016-4 and 311504/2017-5; and Fundação de Amparo à Pesquisa do Espírito Santo - Brasil (FAPES) – grant 84412844/2018.

We have developed a self-driving car, named Intelligent Autonomous Robotic Automobile (IARA, Fig. 1), whose autonomy system is based on precise localization [15]. IARA's hardware is composed of a Ford Scape Hybrid retrofitted with an assortment of sensors and processing units. Its software is composed of many modules, which includes a mapper [16], a localizer [17], a path planner, a behavior selector, a motion planner [18], an obstacle avoider [19], and a controller [20]. Fig. 2 shows a front view of IARA with some of the various sensors visible. Using an environment map, the current pose of IARA in this map, a final goal in this map, and a road network that contains the final goal, the IARA's path planner computes a series of waypoints (Fig. 1) that define the path that IARA has to follow in the time horizon of the future 10 seconds, approximately [15]. Thus, it is possible to anticipate IARA's path and, using it, to project relevant points of required attention in the camera images (Fig. 1).

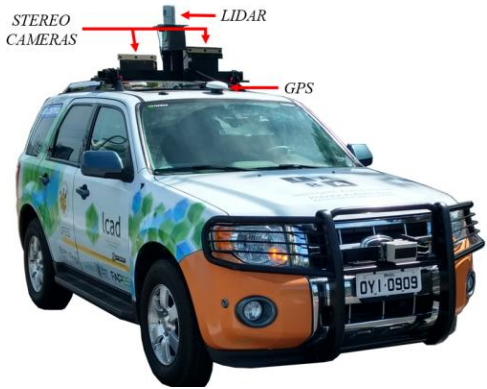

Fig. 2 – Intelligent and Autonomous Robotic Automobile (IARA). Red arrows highlight IARA's sensors that are used to build an internal representation of the external world. A video of IARA operating in autonomous mode is available at https://youtu.be/o_NU23fpZhw.

In this paper, we present a bio-inspired foveated technique for long-range vehicles detection that uses a deep convolutional neural network (DCNN) for vehicles detection in different image scales. The camera used for detection is installed on the roof of IARA. The DCNN receives as input (i) the original image and (ii) crops of the original image that are obtained using waypoints computed by IARA's path planner. The same vehicle may be detected in different crops of the same image, which may be interpreted as more than a single vehicle by the system. To solve this problem, we apply an overlap filter to discard or keep detections, if they correspond or not to the same vehicle. Borders of crops do not need special attention, since the crops taken are large enough to cover the complete region of interest in each point of the original image (see Fig. 1). The proposed *augmented-range vehicle detection system* (ARVDS) increases the Average Precision (AP) of detection from 29.51% (using a single whole image) to 63.15%.

This paper is organized as follows. After this introduction, in Section II we discuss related works. In Section III, we present an overview of the proposed system. In Section IV, we detail the experimental methodology employed to test the proposed system, while, in Section V, we present experimental results. We close, in Section VI, with our conclusions and directions for future work.

## II. RELATED WORK

Several techniques for long-range object detection were proposed in the literature. Zhang et al. [21] propose a street scene layout estimation method for distant object detection. The scene layout estimation method is based on a conditional random field (CRF) model on the layout of the sky and ground boundaries in the images of the environment. With the inferred scene layout, they exploit the scene geometry to zoon and enhance the image in the regions of interest. They, then, use a deformable part model (DPM [22]) for car detection in these regions. We use the self-driving car future path, computed by its path planner, for selecting regions of interest and a DCNN for car detection (DCNNs have shown significantly better performance than DPMs on object detection [23]).

Batzer et al. [24] present an approach for generating hypotheses of small or distant objects in images based on a voting scheme. The environment is modeled by very few, large regions with homogeneous appearance, which are computed using image statistics extracted from the image only, without a priori knowledge about the environment. Small areas that cannot be assigned to one of these regions due to the variance of the pixels intensity are potential object candidates. They do not detect objects, though, but only classify image pixels as possibly belonging to an object of interest or not.

Ohn-Bar et al. [25] propose a multi-scale structure (MSS) approach for investigating general object detection with a CNN at multiple image scales. Instead of training and testing the CNN over local image regions (either a sliding window or region proposals), the MSS approach operates on all scales of an image pyramid in training and testing. This approach has a performance significantly inferior than the YOLOv2 CNN, which we use in this work. For example, on the PASCAL VOC 2007 dataset MSS achieves a mAP (mean Average Precision) of 42.74%, while YOLOv2 achieves a mAP of 78.6%.

LaLonde et al. [26] present a two-stage spatio-temporal CNN, named ClusterNet, that combines appearance and motion information for object detection in wide area motion imagery (WAMI). The first stage takes in a set of large video frames, combines the motion and appearance information within the CNN, and proposes regions of objects of interest (ROOBI). The second stage (FoveaNet) estimates the location of all objects in the ROOBI simultaneously via heatmap estimation. Experiments were executed on the WPAFB 2009 dataset, which has almost 2.4 million vehicle detections spread across only 1,025 frames of video, averaging over 2,000 vehicles to detect in every frame – a truly WAMI dataset. We are interested in vehicles detection in the context of self-driving cars operation, however. The proposed ClusterNet was designed for the context of WAMI and its performance in the context of self-driving cars is yet to be examined.

## III. AUGMENTED-RANGE VEHICLE DETECTION SYSTEM (ARVDS)

The *augmented-range vehicle detection system* (ARVDS) works as follows (see Fig. 3). During the operation of the self-driving car, we periodically capture (i) an image using IARA's front-facing camera, (ii) IARA's current position, and (iii) the waypoints of the next 150 m of the current plan produced by IARA's path planner. We then project these waypoints into the camera's image (Fig. 3 – waypoints projection) and take crops of the image associated with the different waypoints (Fig. 3 – image crop). After that, we send the original image and its crops to a neural network trained for detecting cars (Fig. 3 – car detection). Finally, we summarize the neural network detections into a single set of detections (Fig. 3 – detection filtering) by removing eventual duplicates (detections of the same car in multiple crops). In the following subsections, we describe all this process in more detail.

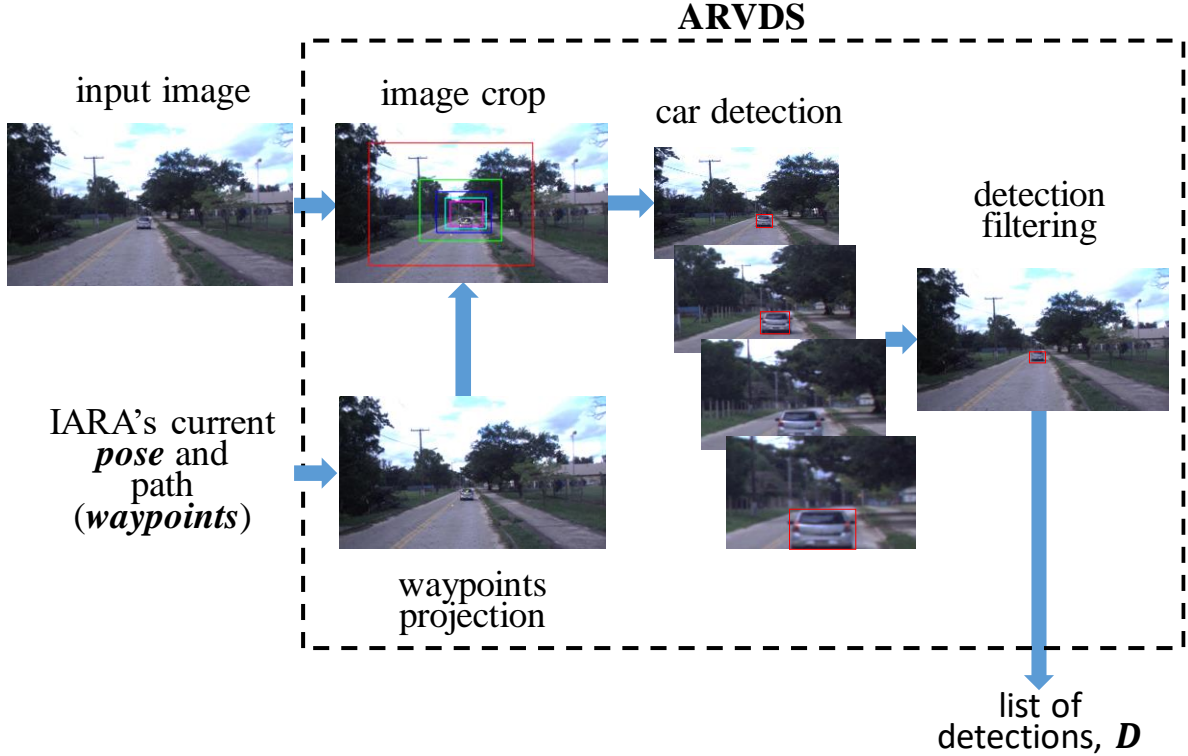

Fig. 3 - System Overview.

### A. Waypoints Projection

Self-driving cars must anticipate the path they will follow throughout a process known in the literature as path planning (or motion planning) [27]. To establish a path, the path planner used in IARA (the self-driving car employed in the research) defines a list of 0.5 m-spaced waypoints extending 150 m in front of IARA [15]. In the waypoints projection phase (Fig. 3 – waypoints projection) of ARVDS, we project the waypoints of IARA's current path onto the current image of the IARA's front-facing camera. To explain how we do that, we are going to use the following notation.

**Notation:** We write scalars in lowercase italic letters ($a$), vectors in bold lowercase ($\mathbf{a}$), matrices using boldface capitals ($\mathbf{A}$), and lists in boldface italic capitals ($\boldsymbol{A}$). Scalars and vectors may have subscripts and/or superscripts to better characterize them (i.e., $a_x$, or $\mathbf{a}_x^j$). A 3D rigid-body transformation matrix that takes points from coordinate system $a$ to coordinate system $b$ will be denoted by $\mathbf{T}_a^b$, with $\mathbf{T}$ for 'transformation'. We have 4 coordinate systems: world, car, camera, and sensor board (in IARA, the sensors are mounted in a sensor board). They are all oriented as follows: $x$ = forward, $y$ = left, $z$ = up.

In the IARA autonomous system, a path, $\boldsymbol{L} = \{\mathbf{w}_1^w, \ldots, \mathbf{w}_j^w, \ldots \mathbf{w}_{|L|}^w\}$, is a list of waypoints, $\mathbf{w}_j^w = (w_{j_x}^w, w_{j_y}^w, w_{j_z}^w)$, which define positions in *world* in the Universal Transverse Mercator (UTM) coordinate system. Given the IARA's current UTM pose in the world, $\mathbf{p}^w = (p_x^w, p_y^w, p_z^w)$, we can project a waypoint, $\mathbf{w}_j^w$, of $\boldsymbol{L}$ to a point $\mathbf{w}_j^c$ in the IARA's front *camera* coordinate system using Equation (1),

$$\mathbf{w}_j^c = \mathbf{T}_b^c \mathbf{T}_p^b \mathbf{T}_w^p \mathbf{w}_j^w \qquad (1)$$

where $\mathbf{T}_w^p$ is the transformation matrix that takes a point from world coordinates to the IARA's *pose* coordinates, $\mathbf{T}_p^b$ is the transformation matrix that takes a point from the IARA's pose coordinates to the sensor *board* coordinates, and $\mathbf{T}_b^c$ is the transformation matrix that takes a point from the sensor board coordinates to the camera coordinates (Fig. 4). The point $\mathbf{w}_j^c = (w_{j_x}^c, w_{j_y}^c, w_{j_z}^c)$ is projected to a point $\mathbf{w}_j^i = (w_{j_u}^i, w_{j_v}^i)$ in the camera *image* plane (see Fig. 5) using equations (2) and (3) [28]:

$$w_{j_u}^i = \frac{f_x}{s} \frac{w_{j_y}^c}{w_{j_x}^c} + k_u \qquad (2)$$

$$w_{j_v}^i = \frac{f_y}{s} \frac{-w_{j_z}^c}{w_{j_x}^c} + k_v \qquad (3)$$

where $f_x$ and $f_y$ are the camera's focal lengths in meters, $s$ is the pixel size in meters, and $k_u$ and $k_v$ are the coordinates of the camera's principal point in pixels.

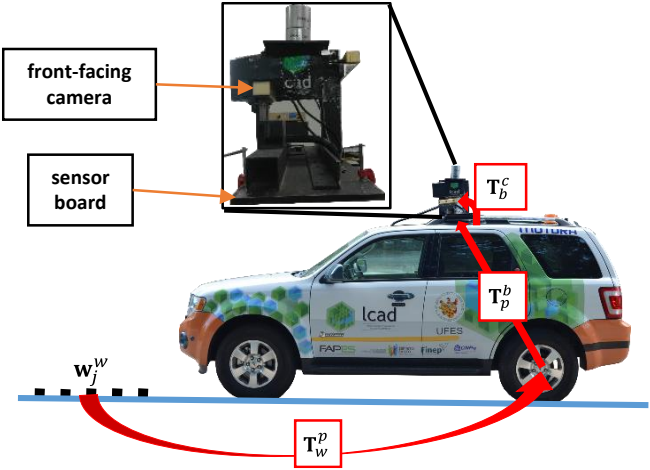

Fig. 4. Waypoints projection. The black dots represent the waypoints of IARA's current path, coded in world coordinates that, in the IARA autonomous system, follows the Universal Transverse Mercator (UTM) coordinate system. The $\mathbf{T}_w^p$ transformation matrix transforms a point in world coordinates into a point in IARA's pose coordinates. The $\mathbf{T}_p^b$ transformation matrix transforms a point in IARA's pose coordinates into a point in sensor board coordinates. Finally, the $\mathbf{T}_b^c$ transformation matrix transforms a point in sensor board coordinates into a point in camera coordinates.

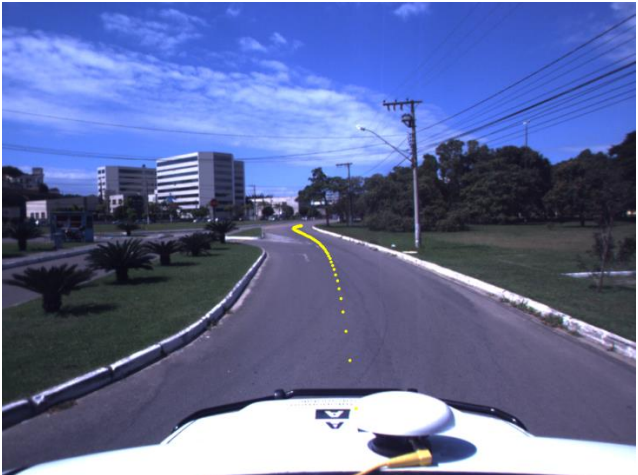

Fig. 5. Waypoints projection. The yellow points are waypoints of IARA's current path projected into the image plane.

### B. Image Crop

In the image crop phase (Fig. 3 – image crop) of ARVDS, we take crops of the image of IARA's front-facing camera aligned with the projection of waypoints of the IARA's path, $L$, into this image. As these waypoints are very close together in world (0.5 m-spaced, see Fig. 5), we select only a number, $n$, of them as reference for $n$ image crops. These $n$ waypoints are equally spaced of $d$ meters (Fig. 6), with the first localized $d$ meters ahead of the current front-camera pose in the world, $\mathbf{i}^w$, given by Equation (4),

$$\mathbf{i}^w = \mathbf{T}_b^c \mathbf{T}_p^b \mathbf{p}^w. \qquad (4)$$

A crop, $\mathbf{c}_j$, is defined by its width, $c_{j_w}$, height, $c_{j_h}$, and its top left coordinates in the current image, $c_{j_u}$ and $c_{j_v}$. The further away the waypoint is from IARA, the smaller is the size of the image crop (Fig. 6). The first crop has an *ad hoc* defined size of 60% of the IARA's front-facing camera image size, while the remaining crops' sizes are halved at every $d$ meters, i.e.,

$$c_{j_w} = \frac{0.6 * m_w}{j}, j = 1, 2, \dots, n, \qquad (5)$$

$$c_{j_h} = \frac{0.6 * m_h}{j}, j = 1, 2, \dots, n, \qquad (6)$$

where $m_w$ is IARA's front-facing camera image width, and $m_h$ is IARA's front-facing camera image height. The top left coordinates of $\mathbf{c}_j$ in IARA's front-facing camera image are

$$c_{j_u} = w_{j_u}^i - \frac{c_{j_w}}{2}, j = 1, 2, \dots, n, \text{ and} \qquad (7)$$

$$c_{j_v} = w_{j_v}^i - \frac{c_{j_h}}{2} * 1.5, \ j = 1, 2, \dots, n, \qquad (8)$$

i.e., each crop, $\mathbf{c}_j$, is horizontally centered in the position of the waypoint $\mathbf{w}_j^i$ in the image, and slightly moved up (i.e., vertically) with respect to the position of $\mathbf{w}_j^i$ in the image (see Fig. 6).

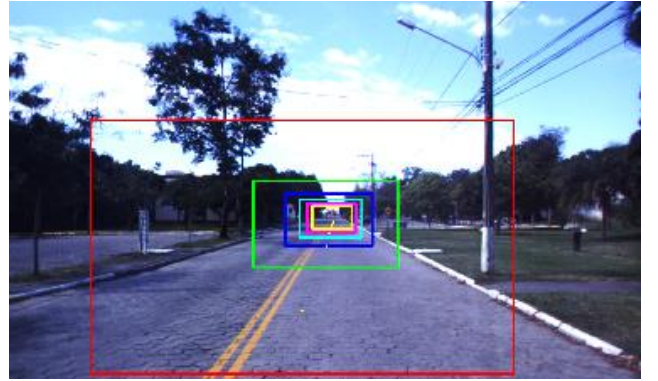

Fig. 6. Waypoints equally spaced of $d$ meters and corresponding image crops. Each colored rectangle represents a crop.

### C. Car Detection

In the car detection phase (Fig. 3 – image crop) of ARVDS, we employ the deep convolutional neural network (DCNN) named You Only Look Once version 2 (YOLOv2) [12] for detecting cars in the image of the front-facing camera and its crops. YOLOv2 is a state-of-the-art DCNN that, on a NVIDIA Titan X GPU, processes images at 40-90 fps and has a mAP on VOC 2007 [29] of 78.6% and a mAP of 48.1% on COCO test-dev [30]. These results obtained by YOLOv2 overcame those reached by other state-of-art detection systems, which motivated us to employ it in this work [12].

Table 1 presents the architecture of YOLOv2. It is basically composed of many convolutional layers with padding, some max pooling layers, and some *ad hoc* layers [12]. YOLOv2 was implemented in Darknet, an open source neural network framework written in C and CUDA [31].

Table 1: YOLOv2 architecture

| Nº | Type | Filters | Size/Stride | Output |
|---|---|---|---|---|
| 1 | Convolutional | 32 | 3×3 | 608×608 |
| 2 | Maxpool | | 2×2/2 | 304×304 |
| 3 | Convolutional | 64 | 3×3 | 304×304 |
| 4 | Maxpool | | 2×2/2 | 152×152 |
| 5 | Convolutional | 128 | 3×3 | 152×152 |
| 6 | Convolutional | 64 | 1×1 | 152×152 |
| 7 | Convolutional | 128 | 3×3 | 152×152 |
| 8 | Maxpool | | 2×2/2 | 76×76 |
| 9 | Convolutional | 256 | 3×3 | 76×76 |
| 10 | Convolutional | 128 | 1×1 | 76×76 |
| 11 | Convolutional | 256 | 3×3 | 76×76 |
| 12 | Maxpool | | 2×2/2 | 38×38 |
| 13 | Convolutional | 512 | 3×3 | 38×38 |
| 14 | Convolutional | 256 | 1×1 | 38×38 |
| 15 | Convolutional | 512 | 3×3 | 38×38 |
| 16 | Convolutional | 256 | 3×3 | 38×38 |
| 17 | Convolutional | 512 | 1×1 | 38×38 |
| 18 | Maxpool | | 2×2/2 | 19×19 |
| 19 | Convolutional | 1024 | 3×3 | 19×19 |
| 20 | Convolutional | 512 | 1×1 | 19×19 |
| 21 | Convolutional | 1024 | 3×3 | 19×19 |
| 22 | Convolutional | 512 | 1×1 | 19×19 |
| 23 | Convolutional | 1024 | 3×3 | 19×19 |
| 24 | Convolutional | 1024 | 3×3 | 19×19 |
| 25 | Convolutional | 1024 | 3×3 | 19×19 |
| 26 | Route | 17 | | |
| 27 | Convolutional | 64 | 1×1 | 38×38 |
| 28 | Reorg | | /2 | 19×19 |
| 29 | Route | 28,25 | | |
| 30 | Convolutional | 1024 | 3×3 | 19×19 |
| 31 | Convolutional | 425 | 1×1 | 19×19 |
| 32 | Detection | | | |

In ARVDS, YOLOv2 receives as input the image of the front-facing camera and its crops, and returns as output a bounding box for each object detected in them and the corresponding object class. YOLOv2 detects 80 different classes, such as "person", "car", "aeroplane", "boat", "bird", and "dog". Currently, in ARVDS, only bounding boxes classified as "car" are considered. We made this choice to better evaluate its performance, since our evaluation dataset consider only objects of class "car". However, ARVDS can be easily modified to consider any object of interest that can be detected by the DCNN.

### D. Detection Filtering

In the detection filtering phase of ARVDS (Fig. 3 – detection filtering), we put together into a matrix all bounding boxes returned by YOLOv2. The lines of this matrix correspond to each image sent to YOLOv2, and the columns contain the bounding boxes and corresponding classes of objects detected. The same car may be detected multiple times in different crops of the same image (Fig. 7(a)), which may be interpreted as more than a single car by the ARVDS. To solve this problem, we employ an overlap filter to discard or keep bounding boxes (Fig. 7(b)), if they correspond or not to the same car.

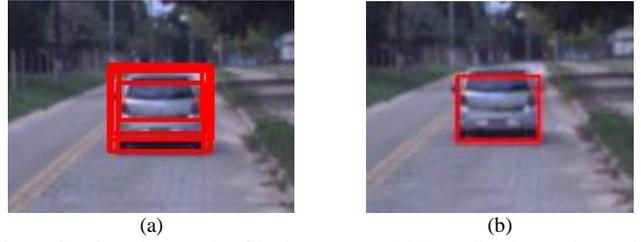

Fig. 7. Car detection overlap filtering process. (a) Bounding boxes detected in different crops shown together in the original image. (b) Result of our bounding box filtering process.

The overlap filter works as follows. If one of the input images of YOLOv2 contains any detection, the bounding boxes of these detections are stored in the detection matrix, **D**. Then, the filter examines each line of **D** in sequence, starting with the line of the original image, and proceeding with the lines of the crops, from the largest to the smallest. All bounding boxes of the first line of **D** that contains any detection are added to the list of detections, ***D***. After that, for each bounding box encountered in the remaining lines of **D**, we check if there is an overlap of this bounding box with any of the bounding boxes in ***D***. We consider that there was an overlap if the intersection over union (IoU [32]) is larger than 50%. If there is an overlap, the bounding box is not added to ***D***. The output of ARVDS is the list ***D***.

### IV. EXPERIMENTAL METHODOLOGY

To evaluate the performance of ARVDS, we used the hardware and software infrastructure available in the IARA self-driving car (Fig. 2) – this made the implementation and experimental evaluation of ARVDS significantly easier. Using IARA, we captured thousands of images of real traffic situations and annotated these images in order to build an evaluation dataset. Finally, using proper metrics, we carried out experiments to examine the advantages of using ARVDS for augmented range vehicle detection.

### A. Hardware Setup

IARA is a robotic car platform based on a Ford Escape Hybrid, which was modified to allow electronic control of steering, throttle, brakes, gears, and several signalization items; and to provide the car odometry and power supply for computers and sensors. The power is taken from the hybrid-system battery and converted from 330V DC to 120V AC. IARA's computer is a Dell Precision R5500 with 2 Xeon X5690 six-core 3.4 GHz processors, and one NVIDIA TITAN Xp. The IARA's sensors include one Velodyne HDL 32-E LIDAR, one Trimble RTK GPS, one Xsens MTi IMU, and one Bumblebee XB3 stereo camera.

To find the values of the terms of the transforms between the different systems of coordinates of IARA ($\mathbf{T}_b^c$, $\mathbf{T}_p^b$, and $\mathbf{T}_w^p$), we manually calibrated the position of the sensors and sensor board using several in-house tools. The constants in equations (2) and (3) were provided by PointGrey (we used rectified camera images).

## B. Software

IARA's software is based on the Carnegie Mellon Robot Navigation Toolkit (CARMEN [33]), which is a modular open source software collection for mobile robot control. The Computational Intelligence Research Group at the High-Performance Computing Laboratory (*Laboratório de Computação de Alto Desempenho* – LCAD, www.lcad.inf.ufes.br) of the Federal University of Espírito Santo (*Universidade Federal do Espírito Santo* – UFES, Brazil) has significantly extended and currently maintain a version of CARMEN, available at https://github.com/LCAD-UFES/carmen_lcad. For detailed information about IARA's modules, readers are referred to Badue et al. [15].

## C. Evaluation Dataset

To test our system, we logged all the IARA's sensor data in different traffic scenarios. The evaluation dataset was extracted from the logged data with a camera frame rate of 15 fps. The dataset is composed of 1,085 images of 1,280×768-pixel. The vehicles of interest (cars) were manually annotated in dataset's images. The database and the annotations are available at https://goo.gl/5gxwEM.

## D. YOLOv2

The YOLOv2 DCNN was used without fine-tuning, i.e., it was used with original weights obtained from training with ImageNet [34] and Common Objects in Context (COCO) [30] datasets. The input size employed was 608×608.

## E. Metrics

To evaluate the performance of ARVDS, we use the average precision (AP) metric. The formulation of AP employed is the same used in the PASCAL VOC 2010 challenge [29] and is defined as the area under the precision-recall curve obtained with a fixed intersection over union (IoU) threshold (0.5 in this work).

## V. RESULTS

The experiments carried out to test ARVDS were performed as follows. We used IARA's software infrastructure (Section IV.B) to play the same logs of sensor data employed to build our evaluation dataset (Section IV.C), so that the ARVDS would run as if it was operating connected to IARA in a real world scenario. While the logs were being played, we saved the output of ARVDS (the list of detections, *D*; Section III.D). We then compared the saved output of ARVDS with the ground truth in our evaluation dataset.

We executed six experiments as described above, varying the number of crops taken from the input image and sent to the car detection phase of ARVDS (Fig. 3) from zero to 5, and measured the average precision (AP) of the ARVDS detections considering the ground truth of our dataset. The results of these six experiments are shown in Fig. 8

As Fig. 8 shows, the performance of ARVDS increases as the number of crops increases. With no crops, the performance is the same as that of a system that employs YOLOv2 only. In this case, the AP is 29.51%. However, as the number of crops increases from 1 to 5, the performance increases from 35.54% to 63.15%. The performance saturates after 4 crops, though. This is to be expected, since the size of the forth crop is only 192×115-pixel. Fig. 9 presents the precision×recall curves and the AP (or the approximate *area under the curves*, AUCs, i.e., the AP employed in the PASCAL VOC 2010 challenge [29]) for different numbers of crops.

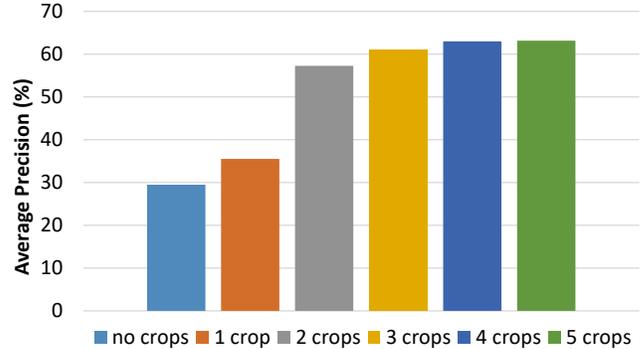

Fig. 8. Average precision of detections for different numbers of crops.

Fig. 10 illustrates how far ARVDS can detect cars. In Fig. 10(a) we show an input image of ARVDS, which has 1,280×768-pixel, while in Fig. 10(b) we show the fourth crop and Fig. 10(c) the fifth crop (153×92-pixel). The cars detected in the ARVDS fifth crop are almost invisible in the input image. A video of the ARVDS working can be found at: https://goo.gl/5gxwEM.

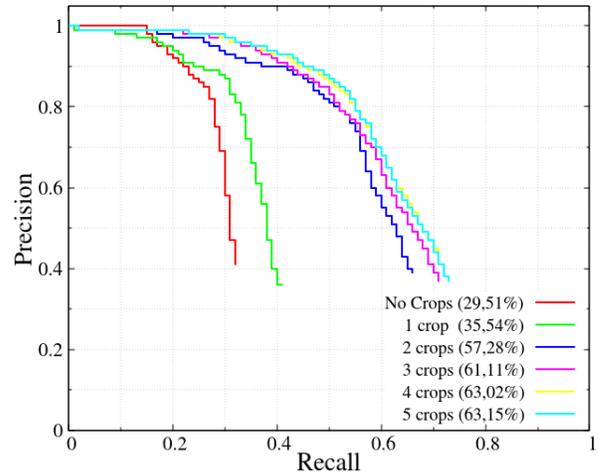

Fig. 9. Precision×Recall curves and AP for different numbers of crops.

## VI. CONCLUSION AND FUTURE WORK

In this paper, we presented the augmented-range vehicle detection system (ARVDS). ARVDS was designed for self-driving cars and uses a deep convolutional neural network (DCNN) for vehicles detection in different image scales. For that, it employs a bio-inspired foveated technique where the DCNN receives as input (i) the image captured by a front-facing camera in the self-driving car, and (ii) crops of this image obtained using waypoints computed by the self-driving car's path planner. That is, we take crops of the image aligned with the projection of waypoints of the self-driving car's future path in the image in such a way as to emulate the way we look ahead while we are driving. These crops are enlarged and sent to the DCNN, which, focusing on them, is able to detect vehicles at a large distance expending much less processing power than would be necessary if it had to examine the whole input image in several scales.

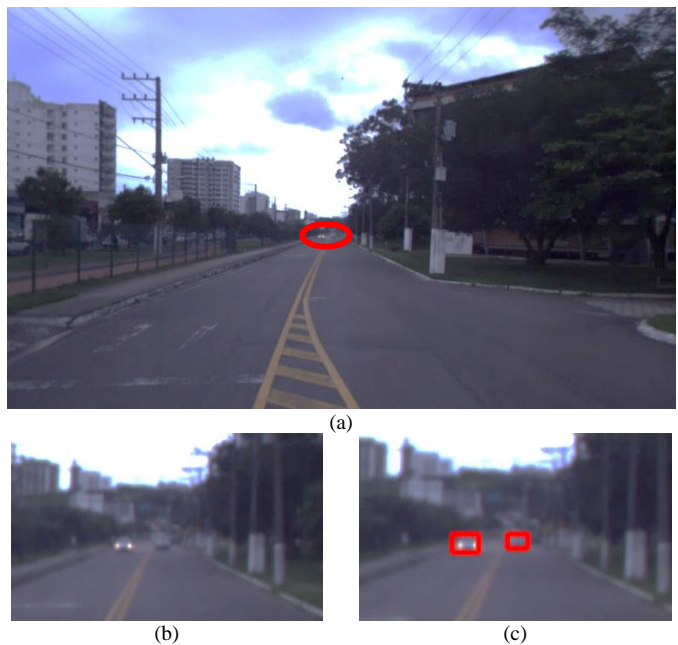

Fig. 10. Distant cars detected by ARVDS.

We evaluated ARVDS in real world scenarios using the IARA self-driving car hardware and software, and an annotated evaluation dataset. Our results show that ARVDS can improve the average precision performance of the DCNN from 29.51% to 63.15% in the examined scenarios. These significant performance gains obtained by ARVDS indicate that it would be helpful for augmented-range detections in self-driving cars.

As directions for future work, we plan to examine the benefits of employing better cameras and more advanced neural networks to improve ARVDS performance. Also, we plan to fuse LiDAR data with the camera's images to allow precise estimation of the pose of the detected vehicles with respect to that of the self-driving car.

ACKNOWLEDGMENT

The authors would like to thank the NVIDIA Corporation for their donation of GPUs.